\documentclass{article}
\usepackage{spconf,amsmath,graphicx}
\usepackage{flushend}
\usepackage{algorithm}
\usepackage{algorithmicx}
\usepackage{mathtools}
\usepackage{algpseudocode}
\usepackage{subfigure}
\usepackage{multirow}
\usepackage{longtable}
\usepackage{amsthm}
\usepackage{amssymb }
\usepackage{amsmath}
\usepackage{bm}
\usepackage{mathrsfs}
\usepackage{amsfonts}
\usepackage{longtable}
\usepackage{multirow}
\usepackage{cite}
\usepackage{xcolor}
\usepackage{graphicx}
\graphicspath{ {./images/} }

\def\W{\mathit{W}}
\def\V{\bm{V}}
\def\Q{\bm{Q}}
\def\K{\bm{K}}
\def\Z{\bm{Z}}

\def\E{\bm{E}}

\def\N{N}
\def\p{p}
\def\R{\mathbb{R}}
\def\Nd{^{\N\times\d}}
\def\d{d}

\def\dh{d_h}

\def\h{h}

\def\j{_{j}}
\def\x{{\mathbf x}}
\def\z{{\mathbf z}}
\def\xp{\x^{\p}}

\def\yb{\mathbf{y}}

\title{ViT-CAT: Parallel Vision Transformers with Cross Attention Fusion for Popularity Prediction in MEC Networks}
\ninept

\name{Zohreh HajiAkhondi-Meybodi$^\dag$, Arash Mohammadi$^\ddag$, Ming Hou$^{\dag\ddag}$, \thanks{This Project was partially supported by Department of National Defence's Innovation for Defence Excellence \& Security (IDEaS), Canada.}\vspace{-.2in}}
\address{\textit{Jamshid Abouei$^{\dag\dag}$, Konstantinos N. Plataniotis$^{\ddag\ddag}$}\\\\
$^\dag$~Electrical and Computer Engineering, Concordia University, Montreal, Canada\\
$^\ddag$~Concordia Institute of Information Systems Engineering (CIISE), Montreal, Canada\\
$^{\dag\ddag}$~Defence Research and Development Canada (DRDC), Toronto, Canada\\
$^{\dag\dag}$~Department of Electrical Engineering, Yazd University, Yazd, Iran\\
$^{\ddag\ddag}$~Electrical and Computer Engineering, University of Toronto, Toronto, Canada}

\begin{document}

\maketitle

\begin{abstract}
Mobile Edge Caching (MEC) is a revolutionary technology for the Sixth Generation ($6$G) of wireless networks with the promise to significantly reduce users' latency via offering storage capacities at the edge of the network. The efficiency of the MEC network, however, critically depends on its ability to dynamically predict/update the storage of caching nodes with the top-$K$ popular contents. Conventional statistical caching schemes are not robust to the time-variant nature of the underlying pattern of content requests, resulting in a surge of interest in using Deep Neural Networks (DNNs) for time-series popularity prediction in MEC networks. 
However, existing DNN models within the context of MEC fail to simultaneously capture both temporal correlations of historical request patterns and the dependencies between multiple contents. This necessitates an urgent quest to develop and design a new and innovative popularity prediction architecture to tackle this critical challenge. The paper addresses this gap by proposing a novel hybrid caching framework based on the attention mechanism. Referred to as the parallel Vision Transformers with Cross Attention (ViT-CAT) Fusion, the proposed architecture consists of two parallel ViT networks, one for collecting temporal correlation, and the other for capturing dependencies between different contents. Followed by a Cross Attention (CA) module as the Fusion Center (FC), the proposed ViT-CAT is capable of learning the mutual information between temporal and spatial correlations, as well, resulting in improving the classification accuracy, and decreasing the model's complexity about $8$ times. Based on the simulation results, the proposed ViT-CAT architecture outperforms its counterparts across the classification accuracy, complexity, and cache-hit~ratio.
\end{abstract}
\maketitle
\vspace{-.15in}
\section{Introduction} \label{sec:introduction}
\vspace{-.1in}
The phenomenal growth in demand for mobile wireless data services, together with the emergence of advanced Internet of Things (IoT) applications bring new technical challenges to wireless communications. According to Ericsson's mobility report~\cite{Ericsson}, global mobile data traffic is projected to exponentially grow from  $67$ exabytes/month in $2021$ to $282$ exabytes/month in $2027$. To accommodate the huge amount of mobile data traffic, Mobile Edge Caching (MEC)~\cite{Hajiakhondi2021, Liu2021, Abbas2018, Khan2020} has emerged as a promising solution for potential deployment in the Sixth Generation ($6$G) of communication networks. MEC networks provide low-latency communication for IoT devices by storing multimedia contents in the storage of nearby caching nodes~\cite{Hajiakhondi2019, Hajiakhondi2020}. The limited storage of caching nodes, however, makes it impossible to preserve all contents on nearby devices. To tackle this challenge, predicting the most popular content is of paramount importance, as it can significantly influence the content availability in the storage of caching nodes and reduce users' latency. 

Existing popularity prediction solutions are typically developed based on statistical models~\cite{Hajiakhondi2019, Hajiakhondi2020, Hajiakhondi2022, Marlin2011, Odic2013}, Machine Learning (ML)-based architectures~\cite{Abidi2020, Ng2019, Kabra2011, Mendez2008}, and  Deep Neural Networks (DNNs)~\cite{Hajiakhondi2021_ICC, Hajiakhondi2022_I0T, Doan2018, Ale2019, Zhang2019, Lin2020, Zhong2020, Wu2019, Wang2019, Mou2019}, among which the latter is the most efficient one for popularity prediction. This is mainly due to the fact that DNN-based models can capture users' interests from raw historical request patterns without any feature engineering or pre-processing. In addition, DNN-based popularity prediction models are not prone to sparsity and cold-start problems with new mobile user/multimedia contents. As a result, recent research has shifted its primary attention to DNN-based frameworks to monitor and forecast the popularity of content using its historical request pattern. A critical aspect of a DNN-based popularity prediction architecture is its ability to accurately capture both temporal and spatial correlations within the time-variant request patterns of multiple contents. While the temporal correlation illustrates the variation of users' preferences over time, spatial correlation reflects the dependency between different multimedia contents. The majority of works in this field~\cite{Ale2019, Zhang2019, Zhong2020, Wu2019}, however, are not appropriately designed to simultaneously capture both dependencies. This necessitates an urgent quest to develop and design a new and innovative popularity prediction architecture, which is the focus of this paper.

\vspace{-.005in}
\noindent
\textbf{Literature Review:} Recently, a variety of promising strategies have been designed to forecast the popularity of multimedia contents with the application to MEC networks. In~\cite{Yu2021}, an auto-encoder architecture was proposed to improve content popularity prediction by learning the latent representation of historical request patterns of contents. To boost the decision-making capabilities of caching strategies, Reinforcement Learning (RL)~\cite{Tang2020, Sadeghi2019} and Convolutional Neural Network (CNN)~\cite{Ndikumana2021}-based caching frameworks were introduced to exploit the contextual information of users. Despite all the benefits that come from the aforementioned works, they relied on a common assumption that the content popularity/historical request patterns of contents would remain unchanged over time, which is not applicable in highly dynamic practical systems. 

To capture the temporal correlation of historical request patterns of contents, several time-series-based DNN caching strategies~\cite{Ale2019, Jiang2020} were introduced, among which Long Short Term Memory (LSTM)~\cite{Zhang2019, Mou2019} is one of the most effective learning models. LSTM, however, suffers from computation/time complexity, unsuitability to capture long-term dependencies, parallel computing, and capturing dependencies between multiple contents. To take into account the correlation among historical request patterns of various contents, a Clustering-based LSTM (C-LSTM) model~\cite{Zhang2021} was proposed to predict the number of content requests in the upcoming time. C-LSTM framework, however, is still prone to computation/time complexity, and parallel computing issues. To tackle the aforementioned challenges, Transformer architectures~\cite{Vaswani2017} have been developed as a time-series learning model, while the sequential data need not be analyzed in the same order, resulting in less training complexity and more parallelization. There has been a recent surge of interest in using Transformers in various applications~\cite{Vaswani2017, Dosovitskiy2019, Gulati2020}. 
The paper aims to further advance this emerging field.

\vspace{-.005in}
\noindent
\textbf{Contribution:}
A crucial aspect of Transformers that has a significant deal of potential for widespread implementation in various Artificial Intelligence (AI) applications is the self-attention mechanism. In our prior work~\cite{Hajiakhondi2021_ICC}, we have shown the superiority of the Vision Transformer (ViT) architecture in comparison to LSTM for the task of predicting the Top-$K$ popular contents. The input of the standard Transformer is a $1$D sequence of token embeddings. To predict the popularity of multiple contents at the same time, we use $2$D images as the input of the ViT network, where each column of the image is associated with the historical request pattern of one content. Generally speaking, $2$D input samples are split into fixed-size patches in the ViT architecture. To help the ViT network to capture not only the temporal correlation of historical request pattern of contents, but also the dependencies between multiple contents, we provide a parallel ViT network, where $2$D input samples split to different types of patches, i.e., time-based patches, and content-time-based patches. Referred to as the parallel Vision Transformers with Cross Attention (ViT-CAT) Fusion, the proposed architecture consists of two parallel ViT networks (one Time-series Self-attention (TS) path, and one Multi-Content self-attention (MC) path), followed by a Cross Attention (CA) as the fusion center. The CA network is used to effectively fuse multi-scale features obtained from TS and MC networks to improve the overall performance of the ViT-CAT classification task. Simulation results based on the real-trace of multimedia requests illustrate that the proposed ViT-CAT architecture outperforms the conventional ViT network and other state-of-the-art counterparts in terms of the cache-hit ratio, classification accuracy, and training complexity. The rest of the paper is organized as follows: Section~\ref{Sec:3} presents the proposed ViT-CAT architecture. Simulation results are presented in Section~\ref{Sec:4}. Finally, Section~\ref{Sec:5} concludes the paper.

\vspace{-.1in}
\section{Proposed ViT-CAT Architecture} \label{Sec:3}
\vspace{-.1in}
In this section, the MovieLens Dataset~\cite{Harper2015} is briefly introduced, followed by describing the dataset pre-processing method. The proposed ViT-CAT popularity prediction model, which is designed to predict the Top-$K$ popular content, will be explained afterward.

\vspace{.05in}
\noindent
\textbf{2.1. Dataset Pre-processing} 

\noindent
MovieLens is a dataset provided by recommender systems, consisting of users' contextual and geographical information, in which commenting on a content is treated as a request~\cite{Zhang2019}. To predict the Top-$K$ popular content, the following four steps are performed to convert MovieLens dataset to $2$D input samples.

\vspace{-.01in}
\noindent
\textit{\textbf{Step 1 - Request Matrix Formation}}: First the dataset is sorted in the ascending order of time for all contents $c_l$, for ($1 \leq l \leq N_c$). Consequently, an ($T \times N_c$) indicator request matrix is generated, where $T$ denotes the total number of timestamps, and $N_c$ is the total number of distinct contents, where $r_{t,l}=1$, if content $c_l$ is requested at time $t$.

\vspace{-.01in}
\noindent
\textit{\textbf{Step 2 - Time Windowing}}: Relying on a common assumption that the most popular contents will be cached during the off-peak time~\cite{Vallero2020}, it is unnecessary to predict the content popularity at each timestamp. Accordingly, we define an ($N_{\mathcal{W}} \times N_c$) window-based request matrix, denoted by $\textbf{R}^{(\mathcal{W})}$, where $N_{\mathcal{W}}=\frac{T}{\mathcal{W}}$ represents the number of time windows with the length of $\mathcal{W}$, where $\mathcal{W}$ is the time interval between two consecutive updating times. For instance, $r^{(w)}_{t_u,l}= \sum \limits_{t=(t_u-1)\mathcal{W}+1}^{t_u\mathcal{W}} r_{t,l}$ represents the total number of requests of content $c_l$ between two updating times $t_u-1$ and $t_u$.

\vspace{-.01in}
\noindent
\textit{\textbf{Step 3 - Data Segmentation}}: To generate $2$D input samples, the window-based request matrix $\textbf{R}^{(\mathcal{W})}$ is segmented via an overlapping sliding window of length $L$. Accordingly, the modified dataset, denoted by $\mathcal{D}=\{(\textbf{X}_u, \textbf{y}_u)\}_{u=1}^{M}$ is prepared, where $M$ is the total number of input samples. Term $\textbf{X}_u \in \mathbb{R}^{ L \times N_c}$ is $2$D input samples, representing the request patterns of all contents before updating time $t_u$ with the length of $L$. Finally, the term $ \textbf{y}_u \in \mathbb{R}^{N_c \times 1}$ represents the corresponding labels, where $\sum \limits_{l=1}^{N_c} {y_u}_{(l)} = K$, with $K$ denoting the storage capacity of caching nodes. Note that ${y_u}_{(l)}=1$ indicates that content $c_l$ would be popular at $t_{u+1}$, otherwise it would be zero. Next, we describe the data labeling method.

\vspace{-.01in}
\noindent
\textit{\textbf{Step 4 - Data Labeling}}: Given the historical request patterns of contents $\textbf{X}_u$ as the input of the ViT-CAT architecture, we label contents as popular and non-popular, according to the following criteria:
\begin{itemize}
\item[$i)$] \textit{Probability of Requesting a Content}: Probability of requesting content $c_l$, for ($1 \leq l \leq N_c$), at updating time $t_u$, is obtained by $p_l^{(t_u)}=\dfrac{r^{(w)}_{l,t_u}}{\sum \limits_{l=1}^{N_c}r^{(w)}_{l,t_u}}$.
%
\item[$ii)$] \textit{Skewness of the Request Pattern}: The skewness of the request pattern of  content $c_l$, for ($1 \leq l \leq N_c$), is denoted by $\zeta_l$, where negative skew indicates the ascending request pattern of content $c_l$ over time.
\end{itemize}
Accordingly, the Top-$K$ popular contents will be labeled with $y_{u,l}=1$, where $c_l$ for ($1 \leq l \leq K$) has negative skew and the highest probability. This completes the presentation of data preparation. Next, we present the proposed ViT-CAT architecture.

\begin{figure*}[t!]
\centering
\includegraphics[scale=0.3]{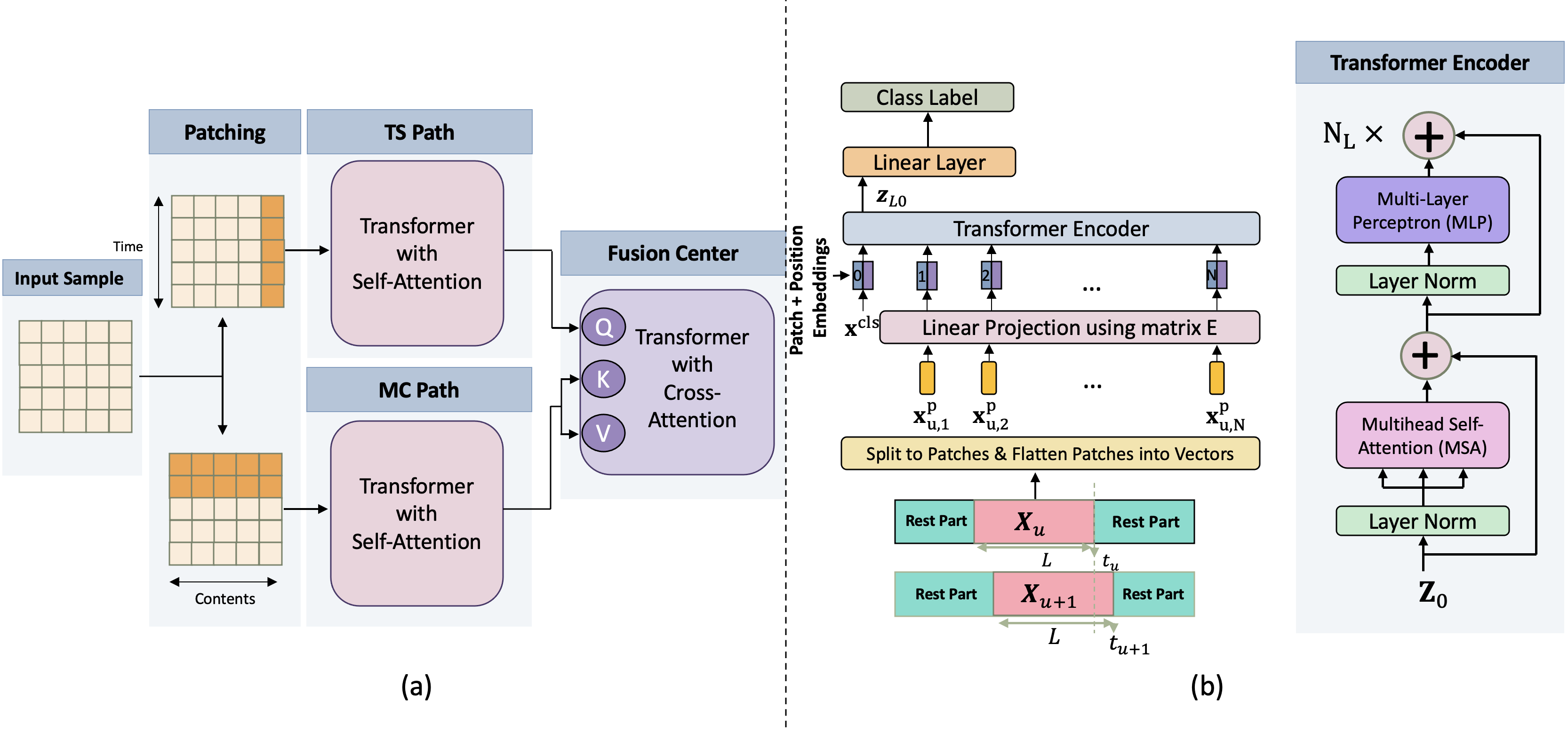}
\vspace{-.1in}
\caption{\footnotesize (a) Block diagram of the proposed ViT-CAT architecture, (b) Pipeline of the ViT architecture, and the Transformer encoder.}\label{BlockDiagram}
\vspace{-.1in}
\end{figure*}


\vspace{.05in}
\noindent
\textbf{2.2. ViT-CAT Architecture}

\noindent
In this subsection, we present different components of the proposed ViT-CAT architecture, which is developed based on the attention mechanism. As shown in Fig.~\ref{BlockDiagram}(a), the ViT-CAT architecture consists of two parallel paths, named Time-Series (TS)-path and Multi-Content (MC)-path, performed based on self-attention mechanism, followed by a Cross-Attention (CA) module as the fusion layer.

\vspace{.05in}
\noindent
\textit{\textbf{A. Patching}} 

\noindent
Generally speaking, the input of the Transformer encoder in the ViT network is a sequence of embedded patches, consisting of patch embedding and positional embedding. In this regard, the $2$D input samples $\textbf{X}_u$ is split into $N$ non-overlapping patches, denoted by $\textbf{X}_{u,p} = \{\textbf{x}_{u,p}^{i}\}_{i=1}^{N}$. As can be seen from Fig.~\ref{BlockDiagram}(a), we apply two following patching methods for TS-path and MC-path:

\begin{itemize}
\item[$i)$] \textit{Time-based Patching}: To capture the temporal correlation of contents, we use time-based patching for the TS-path, where the size of each patch is  $S= L\times 1$. More precisely, time-based patching separately focuses on the request pattern of each content for a time sequence with a length of $L$, where the total number of patches is $N = (L \times N_c) /(L\times 1)= N_c $, which is the total number of contents. 
\item[$ii)$] \textit{Content and Time-based Patching}: In the MC-path, the main objective is to capture the dependency between all $N_c$ contents for a short time horizon $T_s$, therefore, we set the size of each patch to $S = T_s \times N_c$, where the number of patches is $N = (L \times N_c) /(T_s\times N_c)= L/T_s$, with $T_s \ll L$.
\end{itemize}

\noindent
\textit{\textbf{B. Embedding}} 

\noindent
Each patch is flattened into a vector $\xp_{u,\j} \in \R^{N \times S }$ for ($1 \leq j \leq \N$). Referred to as the patch embedding, a linear projection  $\E\in\R^{S\times d}$ is used to embed vector $\xp_{u,\j}$ into the model's dimension $\d$. Then, a learnable embedding token $\x^{cls}$ is added to the beginning of the embedded patches. Finally, to encode the order of the input sequences, the position embedding $\E^{pos}\in\R^{(\N + 1)\times\d}$, is appended to the patch embedding, where the final output of the patch and position embeddings, denoted by $\Z_0$, is given by
\begin{eqnarray}
\Z_0 = [\x^{cls}; \xp_{u,1}\E; \xp_{u,2}\E;\dots; \xp_{u,\N}\E] + \E^{pos}. \label{eq:patch}
\end{eqnarray}

\noindent
\textit{\textbf{C. Transformer Encoder}} 

\noindent
As shown in Fig.~\ref{BlockDiagram}(b), the Multi-Layer Perceptron (MLP) module, consisting of two Linear Layers (LL) with Gaussian Error Linear Unit (GELU) activation function, and the Multihead Self-Attention (MSA) block, are two modules that comprise each layer of the Transformer encoder~\cite{Vaswani2017}, where the total number of layers is denoted by $N_L$. Given the sequence of vectors $\Z_0$ as the input of the Transformer encoder, the outputs of the MSA and MLP modules of layer $\mathcal{L}$, for ($1 \leq \mathcal{L} \leq N_L$), are represented by 
\begin{eqnarray}
\Z^{'}_{\mathcal{L}} &=& MSA(LayerNorm(\Z_{\mathcal{L}-1})) + \Z_{\mathcal{L}-1},\label{eq:MSA}\\
\Z_{\mathcal{L}} &=& MLP(LayerNorm(\Z^{'}_{\mathcal{L}})) + \Z^{'}_{\mathcal{L}}, \label{eq:MLP}
\end{eqnarray}
where the degradation issue is addressed via a layer-normalization. Finally, the output of the Transformer, denoted by $\Z_{\mathcal{L}}^{o}$, is given by
\begin{eqnarray}
\Z_{\mathcal{L}}^{o} = [\z_{\mathcal{L}0}; \z_{\mathcal{L}1}; \dots; \z_{\mathcal{L}\N}],
\end{eqnarray}
where $\z_{\mathcal{L}0}$, which is provided to an LL module, is utilized for the classification task, as follows
\begin{eqnarray}
\yb = LL(LayerNorm(\z_{\mathcal{L}0})).\label{eq:out}
\end{eqnarray}
This completes the description of the Transformer encoder. Next, we briefly explain the SA,  MSA, and CA modules, respectively.

\vspace{.1in}
\noindent
\textit{\textbf{1. Self-Attention (SA):}} To capture the correlation between different parts of the input sample, the SA module is used~\cite{Vaswani2017}, where the input of the SA module is embedded vectors $\Z \in \R\Nd$, where $\Z$ consists of $\N$ vectors with an embedding dimension of $\d$. In this regard, Query $\Q$, Key $\K$, and Value $\V$ matrices with dimension of $\dh$ are defined as
\begin{eqnarray}
[\Q, \K, \V] = \Z\W^{QKV}\label{eq.2},
\end{eqnarray}
where $\W^{QKV} \in \R^{\d\times 3\dh}$ is a trainable weight matrix. After measuring the pairwise similarity between each query and all keys, the output of the SA block $SA(\Z) \in \R^{\N\times \dh}$, which is the weighted sum over all values $\V$, is obtained by
\begin{eqnarray}
SA(\Z) = \text{softmax}(\frac{\Q\K^T}{\sqrt{\dh}}) \V\label{eq.3},
\end{eqnarray}
where the scaled similarity is converted to the probability using $\text{softmax}$, and $\dfrac{\Q\K^T}{\sqrt{\dh}}$ represents the scaled dot-product of $\Q$ and $\K$ by $\sqrt{\dh}$.

\begin{table*}[t]
\centering
\renewcommand\arraystretch{2}
\caption{\footnotesize Variants of the ViT-CAT Architecture.}
\label{table1}
{\begin{tabular}{   c | c c c c c | c | c}
\hline
\hline
\textbf{Model ID}
& \textbf{Layers}
& \textbf{Model dimension $\d$}
& \textbf{MLP layers}
& \textbf{MLP size}
& \textbf{Heads}
& \textbf{Params}
& \textbf{Accuracy}
\\
\hline
\textbf{1}
& 1
& 25
& 1
& 128
& 5
& 201,188
&  84.35 $\%$
\\
\textbf{2}
& 1
& 50
& 1
& 128
& 5
& 435,788
&  94.77$\%$
\\
\textbf{3}
& 1
& 50
& 1
& 128
& 4
& 415,488
& 80.35 $\%$
\\
\textbf{4}
& 1
& 50
& 1
& 64
& 5
& 342,793
& 93.49 $\%$
\\
\textbf{5}
& 1
& 50
& 2
& 64
& 5
& 435,788
&  94.82$\%$
\\
\textbf{6}
& 2
& 50
& 1
& 128
& 5
& 568,185
&  94.84 $\%$
\\
\hline
\end{tabular}}
\vspace{-.2in}
\end{table*}
\vspace{.1in}
\noindent
\textit{\textbf{2. Multihead Self-Attention (MSA):}} The primary objective of the MSA module is to pay attention to input samples from various representation subspaces at multiple spots. More precisely, the MSA module consists of $\h$ heads with different trainable weight matrices $\{\W^{QKV}_i\}^{\h}_{i=1}$, performed $\h$ times in parallel. Finally, the outputs of $\h$ heads are concatenated into a single matrix and multiplied by $\W^{MSA} \in \R^{\h \dh \times \d}$, where $\dh$ is set to $\d / \h$. The output of the MSA module is, therefore, given by
\begin{eqnarray}
MSA(\Z) = [SA_1(\Z); SA_2(\Z); \dots; SA_h(\Z)]W^{MSA}.
\end{eqnarray}

\noindent
\textit{\textbf{3. Cross-Attention (CA):}} The CA module is the same as the SA block, except that the Query $\Q$, Key $\K$, and Value $\V$ are obtained from different input features as shown in Fig.~\ref{BlockDiagram}(a). More precisely, to learn the mutual information between TS and MC paths, the Query $\Q$ comes from the output features of TS-path, while Key $\K$, and Value $\V$ are obtained from the output features of the MC-path. This completes the description of the proposed ViT-CAT architecture. 

\vspace{-.15in}
\section{Simulation Results} \label{Sec:4}
\vspace{-.1in}
In this section, we evaluate the performance of the proposed ViT-CAT architecture through a series of experiments. Given the users' ZIP code in Movielens dataset~\cite{Ndikumana2021}, we assume there are six caching nodes, where the classification accuracy is averaged over all caching nodes. In all experiments, we use the Adam optimizer, where the learning and weight decay are set to $0.001$ and $0.01$, respectively, and binary cross-entropy is used as the loss function for the multi-label classification task. In Transformers, the MLP layers' activation function is ReLU, whereas their output layer's function is~sigmoid.

\noindent
\textbf{Effectiveness of the ViT-CAT Architecture:}
In this subsection, different variants of the proposed ViT-CAT architecture are evaluated to find the best one through trial and error. According to the results in Table~\ref{table1}, increasing the MLP size from $64$ (Model $4$) to $128$ (Model $1$), the model dimension from $25$ to $50$ (Model $1$ to Model $2$), the number of MLP layers from $64$ to $128$ (Models $4$ and $5$), the number of heads from $4$ to $5$ (Models $2$ and $3$), and the number of Transformer layers from $1$ (Model $2$) to $2$ (Model $6$) increase the classification accuracy, while increasing the number of parameters.

\noindent
\textbf{Effect of the Fusion Layer:}
In this experiment, to illustrate the effect of the CA module on classification accuracy, we evaluate the effect of the fusion layer in the proposed ViT-CAT architecture with other baselines, where the parallel ViT architecture in all networks is the same (Model $2$). In this regard, we consider two fusion layers, i.e., the Fully Connected (FL) and the SA layers. According to the results in Table~\ref{table2}, the CA module outperforms the others, since it captures the mutual information between two parallel networks. 

\begin{table}[t]
\centering
\renewcommand\arraystretch{2}
\vspace{-.1in}
\caption{\footnotesize Classification accuracy using different fusion networks.}
\label{table2}
{\begin{tabular}{ | c | c|c|c | }
\hline
\textbf{Model} 
&
\textbf{CA}
&
\textbf{FC}
&
\textbf{SA}
\\
\hline
 \textbf{Accuracy} & 94.77$\%$ &  92.58 $ \%$ &  79.93 $ \%$
\\
\hline
\textbf{Parameters} & 435,788 &  417,171 &  400,788
\\
\hline
\end{tabular}}
\vspace{-.125in}
\end{table}

\begin{figure}[t!]
\centering
\vspace{-.1in}
\includegraphics[scale=0.3]{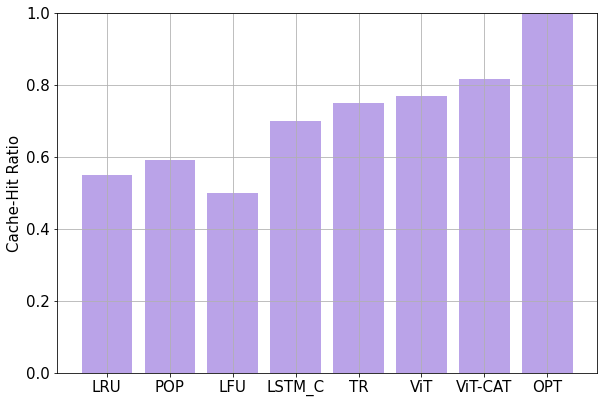}
\vspace{-.1in}
\caption{\footnotesize Comparison with state-of-the-arts based on the cache-hit ratio. }\label{cachehit}
\end{figure}

\noindent
\textbf{Performance Comparisons: }
Finally, we compare the performance of the proposed ViT-CAT architecture in terms of the cache-hit ratio with other state-of-the-art caching strategies, including LSTM-C~\cite{Zhang2019}, TRansformer (TR)~\cite{Nguyen2019}, ViT architecture~\cite{Hajiakhondi2021_ICC}, and some statistical approaches, such as Least Recently Used (LRU), Least Frequently Used (LFU), PopCaching~\cite{Li2016}. With the assumption that the storage capacity of caching nodes is $10\%$ of total contents~\cite{Hajiakhondi2019}, a high cache-hit ratio illustrates that a large number of users' requests are managed through the caching nodes. According to the results in Fig.~\ref{cachehit}, the proposed ViT-CAT architecture outperforms its state-of-the-art counterparts in terms of the cache-hit ratio. As shown in Fig.~\ref{cachehit}, the optimal strategy, which cannot be attained in a  real scenario, is one in which all requests are handled by caching nodes. In addition, we compare the proposed ViT-CAT framework with a single ViT network~\cite{Hajiakhondi2021_ICC}, in terms of accuracy and complexity. It should be noted that the highest accuracy of the ViT-CAT model is $94.84\%$ with $568,185$ number of parameters, while in a single ViT network, the best performance occurs with $93.72\%$ accuracy and $4,044,644$ number of parameters.

\setlength{\textfloatsep}{0pt}

\vspace{-.225in}
\section{Conclusion} \label{Sec:5}
\vspace{-.125in}
In this paper, we presented a parallel Vision Transformers with Cross Attention (ViT-CAT) Fusion architecture to predict the Top-$K$ popular contents in Mobile Edge Caching (MEC) networks. To capture the temporal correlation and the dependency between multiple contents, we employed two parallel ViT networks, followed by a Cross Attention (CA), which was used to learn the mutual information between two networks. Simulation results showed that the proposed ViT-CAT architecture improved the cache-hit ratio, classification accuracy, and complexity when compared to its state-of-the-art.

\end{document}